%% file: main.tex
\documentclass{article}
\let\<\textless
\let\>\textgreater
\PassOptionsToPackage{numbers, compress}{natbib}

\usepackage[final]{nips_2016}

\usepackage[utf8]{inputenc} 
\usepackage[T1]{fontenc}    
\usepackage{hyperref}       
\usepackage{url}            
\usepackage{booktabs}       
\usepackage{amsmath}
\usepackage{amsfonts}       
\usepackage{nicefrac}       
\usepackage{microtype}      
\usepackage{framed}
\usepackage{slashbox}
\usepackage{colortbl}
\usepackage[pdftex]{graphicx}

\DeclareMathOperator*{\argmax}{arg\,max}

\title{Multiresolution Recurrent Neural Networks: \\
An Application to Dialogue Response Generation}

\author{Iulian Vlad Serban${^{*}}{^{\circ}}$ \\ University of Montreal \\ 2920 chemin de la Tour, \\ Montr{\'e}al, QC, Canada
       \And Tim Klinger${^{\diamond}}$ \\ IBM Research \\ T. J. Watson Research Center, \\ Yorktown Heights, \\ NY, USA
       \\\AND Gerald Tesauro${^{\diamond}}$  \\ IBM Research \\ T. J. Watson Research Center, \\ Yorktown Heights, \\ NY, USA
       \And Kartik Talamadupula${^{\diamond}}$  \\ IBM Research \\ T. J. Watson Research Center, \\ Yorktown Heights, \\ NY, USA
       \And Bowen Zhou${^{\diamond}}$  \\ IBM Research \\ T. J. Watson Research Center, \\ Yorktown Heights, \\ NY, USA
       \And Yoshua Bengio${^{\dagger}}{^\circ}$ \\ University of Montreal  \\ 2920 chemin de la Tour, \\ Montr{\'e}al, QC, Canada
       \And Aaron Courville${^{\circ}}$ \\ University of Montreal \\ 2920 chemin de la Tour, \\ Montr{\'e}al, QC, Canada}

\newcommand\blfootnote[1]{%
  \begingroup
  \renewcommand\thefootnote{}\footnote{#1}%
  \addtocounter{footnote}{-1}%
  \endgroup
}

\begin{document}

\blfootnote{* This work was carried out while the first author was at IBM Research.}
\blfootnote{${^{\circ}}$ Email: \{iulian.vlad.serban,yoshua.bengio,aaron.courville\}@umontreal.ca
}
\blfootnote{${^{\diamond}}$ Email: \{tklinger,gtesauro,krtalamad,zhou\}@us.ibm.com}
\blfootnote{${^{\dagger}}$ CIFAR Senior Fellow}

\maketitle
\begin{abstract}
\input{01_abstract}
\end{abstract}

\section{Introduction}
\input{10_introduction}

\section{Model Architecture}
\input{30_model_architecture}

\section{Tasks}
\input{40_tasks}

\section{Coarse Sequence Representations} \label{seq:coarse_seq_rep}
\input{50_coarse_representations}

\section{Experiments}
\input{60_experiments}

\section{Related Work}
\input{20_related_work}

\section{Discussion}
\input{70_discussion}

\subsubsection*{Acknowledgments}

The authors thank Ryan Lowe, Michael Noseworthy, Caglar Gulcehre, Sungjin Ahn, Harm de Vries, Song Feng and On Yi Ching for participating and helping with the human study.
The authors thank Orhan Firat and Caglar Gulcehre for constructive feedback, and thank Ryan Lowe, Nissan Pow and Joelle Pineau for making the Ubuntu Dialogue Corpus available to the public.


\bibliographystyle{natbib}
\begingroup
    \vspace{-2.0mm}
    \footnotesize
    \setlength{\bibsep}{3pt}
    \bibliography{ref}
\endgroup

\small
\newpage
\section*{Appendix}
\input{appendix}



\end{document}

%% file: 01_abstract.tex
We introduce the multiresolution recurrent neural network, which extends the sequence-to-sequence framework to model natural language generation as two parallel discrete stochastic processes: a sequence of high-level coarse tokens, and a sequence of natural language tokens. 
There are many ways to estimate or learn the high-level coarse tokens, but we argue that a simple extraction procedure is sufficient to capture a wealth of high-level discourse semantics.
Such procedure allows training the multiresolution recurrent neural network by maximizing the exact joint log-likelihood over both sequences.
In contrast to the standard log-likelihood objective w.r.t.\@ natural language tokens (word perplexity), optimizing the joint log-likelihood biases the model towards modeling high-level abstractions.
We apply the proposed model to the task of dialogue response generation in two challenging domains:
the Ubuntu technical support domain, and Twitter conversations.
On Ubuntu, the model outperforms competing approaches by a substantial margin, achieving state-of-the-art results according to both automatic evaluation metrics and a human evaluation study.
On Twitter, the model appears to generate more relevant and on-topic responses according to automatic evaluation metrics.
Finally, our experiments demonstrate that the proposed model is more adept at overcoming the sparsity of natural language and is better able to capture long-term structure.
%

%% file: 10_introduction.tex
Recurrent neural networks (RNNs) have been gaining popularity in the machine learning community due to their impressive performance on tasks
such as machine translation \cite{sutskever2014sequence,cho2014learning} and speech recognition~\cite{hinton2012deep}.
These results have spurred a cascade of novel neural network architectures \cite{kumar2015ask}, including attention \cite{bahdanau2014neural,chorowski2015attention}, memory \cite{weston2014memory,graves2014neural,kumar2015ask} and pointer-based mechanisms~\cite{luong2014addressing}. 

The majority of the previous work has focused on developing new neural network architectures within the deterministic sequence-to-sequence framework. In other words, it has focused on changing the parametrization of the deterministic function mapping input sequences to output sequences, trained by maximizing the log-likelihood of the observed output sequence.
Instead, we pursue a complimentary research direction aimed at generalizing the sequence-to-sequence framework to multiple input and output sequences, where each sequence exhibits its own stochastic process.
We propose a new class of RNN models, called multiresolution recurrent neural networks (MrRNNs), which model multiple parallel sequences by factorizing the joint probability over the sequences.
In particular, we impose a hierarchical structure on the sequences, such that information from high-level (abstract) sequences flows to low-level sequences (e.g.\@ natural language sequences).
This architecture exhibits a new objective function for training:
the joint log-likelihood over all observed parallel sequences (as opposed to the log-likelihood over a single sequence), which biases the model towards modeling high-level abstractions.
At test time, the model generates first the high-level sequence and afterwards the natural language sequence.
This hierarchical generation process enables it to model complex output sequences with long-term dependencies.

Researchers have recently observed critical problems applying end-to-end neural network architectures for dialogue response generation~\cite{DBLP:conf/aaai/SerbanSBCP16,li2015diversity}.
The neural networks have been unable to generate meaningful responses taking dialogue context into account,
which indicates that the models have failed to learn useful high-level abstractions of the dialogue.
Motivated by these shortcomings,
we apply the proposed model to the task of dialogue response generation in two challenging domains:
the goal-oriented Ubuntu technical support domain and non-goal-oriented Twitter conversations.
In both domains, the model outperforms competing approaches.
In particular, for Ubuntu, the model outperforms competing approaches by a substantial margin according to both a human evaluation study and automatic evaluation metrics achieving a new state-of-the-art result.



%% file: 30_model_architecture.tex
\subsection{Recurrent Neural Network Language Model}
We start by introducing the well-established recurrent neural network language model (RNNLM) \cite{mikolov2010recurrent,bengio2003neural}. 
RNNLM variants have been applied to diverse sequential tasks, including dialogue modeling~\cite{DBLP:conf/aaai/SerbanSBCP16}, 
speech synthesis~\cite{chung2015recurrent}, handwriting generation~\cite{graves2013generating} and music composition~\cite{boulanger2012modeling}. 
Let $w_1, \dots, w_N$ be a sequence of discrete variables, called tokens (e.g.\@ words), such that $w_n \in V$ for vocabulary $V$.
The RNNLM is a probabilistic generative model,
with parameters $\theta$,
which decomposes the probability over tokens:
\begin{align}
P_{\theta}(w_1, \dots, w_N) = \prod_{n=1}^N P_{\theta}(w_n | w_1, \dots, w_{n-1}).
\end{align}
where the parametrized approximation of the output distribution uses a softmax RNN:
\begin{align}
& P_{\theta}(w_{n+1} = v | w_1, \dots, w_n) = \dfrac{\exp(g(h_n, v))}{\sum_{v' \in V} \exp(g(h_n, v'))}, \\
& h_n = f(h_{n-1}, w_n), \ g(h_n, v) = O_{v}^{\text{T}} h_n,
\end{align}
where $f$ is the hidden state update function, which we will assume is either the LSTM gating unit \cite{hochreiter1997long} or GRU gating unit \cite{cho2014learning} throughout the rest of the paper. For the LSTM gating unit, we consider the hidden state $h_m$ to be the LSTM cell and cell input hidden states concatenated.
The matrix $I \in \mathbb{R}^{d_h \times |V|}$ is the input word embedding matrix, where column $j$ contains the embedding for word index $j$ and $d_h \in \mathbb{N}$ is the word embedding dimensionality. 
Similarly, the matrix $O \in \mathbb{R}^{d_h \times |V|}$ is the output word embedding matrix.
According to the model, the probability of observing a token $w$ at position $n+1$ increases if the context vector $h_n$ has a high dot-product with the word embedding corresponding to token $w$.
Most commonly the model parameters are learned by maximizing the log-likelihood (equivalent to minimizing the cross-entropy) on the training set using gradient descent. 

\subsection{Hierarchical Recurrent Encoder-Decoder}
\label{subseq:hred}

Our work here builds upon that of
Sordoni et al.~\cite{sordoni2015ahier}, who proposed the hierarchical recurrent encoder-decoder model (HRED). 
Their model exploits the hierarchical structure in web queries in order to model a user search session as two hierarchical sequences: a sequence of queries and a sequence of words in each query.
Serban et al.~\cite{DBLP:conf/aaai/SerbanSBCP16} continue in the same direction by proposing to exploit the temporal structure inherent in natural language dialogue. 
Their model decomposes a dialogue into a hierarchical sequence: a sequence of utterances, each of which is a sequence of words.
More specifically, the model consists of three RNN modules: an \textit{encoder} RNN, a \textit{context} RNN and a \textit{decoder} RNN.
A sequence of tokens (e.g.\@ words in an utterance) are encoded into a real-valued vector by the \textit{encoder} RNN.
This in turn is given as input to the \textit{context} RNN, which updates its internal hidden state to reflect all the information up to that point in time.
It then produces a real-valued output vector, which the \textit{decoder} RNN conditions on to generate the next sequence of tokens (next utterance).
Due to space limitations, we refer the reader to \cite{sordoni2015ahier,DBLP:conf/aaai/SerbanSBCP16} for additional information on the model architecture.
The HRED model for modeling structured discrete sequences is appealing for three reasons.
First, it naturally captures the hierarchical structure we want to model in the data. 
Second, the \textit{context} RNN acts like a memory module which can remember things at longer time scales.
Third, the structure makes the objective function more stable w.r.t.\@ the model parameters, and helps propagate the training signal for first-order optimization methods~\cite{sordoni2015ahier}.

\subsection{Multiresolution RNN (MrRNN)}


We consider the problem of generatively modeling multiple parallel sequences.
Each sequence is hierarchical with the top level corresponding to utterances and the bottom level to tokens.
Formally, let $\mathbf{w}_1, \dots, \mathbf{w}_N$ be the first sequence of length $N$ where $\mathbf{w}_n = (w_{n,1}, \dots, w_{n,K_n})$ is the $n$'th constituent sequence consisting of $K_n$ discrete tokens from vocabulary $V^w$.
Similarly, let $\mathbf{z}_1, \dots, \mathbf{z}_N$ be the second sequence, also of length $N$, where $\mathbf{z}_n = (z_{n,1}, \dots, z_{n,L_n})$ is the $n$'th constituent sequence consisting of $L_n$ discrete tokens from vocabulary $V^z$.
In our experiments, each sequence $\mathbf{w}_n$ will consist of the words in a dialogue utterance, and each sequence $\mathbf{z}_n$ will contain the coarse tokens w.r.t.\@ the same utterance (e.g.\@ the nouns in the utterance).

Our aim is to build a probabilistic generative model over all tokens in the constituent sequences $\mathbf{w}_1, \dots, \mathbf{w}_N$ and $\mathbf{z}_1, \dots, \mathbf{z}_N$.
Let $\theta$ be the parameters of the generative model.
We assume that $\mathbf{w}_n$ is independent of $\mathbf{z}_{n'}$ conditioned on $\mathbf{z}_1, \dots, \mathbf{z}_{n}$ for $n' > n$, and factor the probability over sequences:
\begin{align}
P_{\theta}( & \mathbf{w}_1, \dots, \mathbf{w}_N, \mathbf{z}_1, \dots, \mathbf{z}_N) = \prod_{n=1}^N P_{\theta}(\mathbf{z}_n | \mathbf{z}_1, \dots, \mathbf{z}_{n-1}) \prod_{n=1}^N P_{\theta}(\mathbf{w}_n | \mathbf{w}_1, \dots, \mathbf{w}_{n-1}, \mathbf{z}_1, \dots, \mathbf{z}_{n}) \nonumber \\
& =  \prod_{n=1}^N P_{\theta}(\mathbf{z}_n | \mathbf{z}_1, \dots, \mathbf{z}_{n-1}) P_{\theta}(\mathbf{w}_n | \mathbf{w}_1, \dots, \mathbf{w}_{n-1}, \mathbf{z}_1, \dots, \mathbf{z}_{n}),
\end{align}
where we define the conditional probabilities over the tokens in each constituent sequence:
\begin{align}
P_{\theta}(\mathbf{z}_n | \mathbf{z}_1, \dots, \mathbf{z}_{n-1}) & = \prod_{m=1}^{L_n} P_{\theta}(z_{n,m} | z_{n,1}, \dots, z_{n,m-1}, \mathbf{z}_1, \dots, \mathbf{z}_{n-1})  \nonumber \\
P_{\theta}(\mathbf{w}_n | \mathbf{w}_1, \dots, \mathbf{w}_{n-1}, \mathbf{z}_1, \dots, \mathbf{z}_{n}) & = \prod_{m=1}^{K_n} P_{\theta}(w_{n,m} | w_{n,1}, \dots, w_{n,m-1}, \mathbf{w}_1, \dots, \mathbf{w}_{n-1}, \mathbf{z}_1, \dots, \mathbf{z}_{n}) \nonumber
\end{align}

\begin{figure}[ht]
  \centering
  \includegraphics[width=0.825\linewidth]{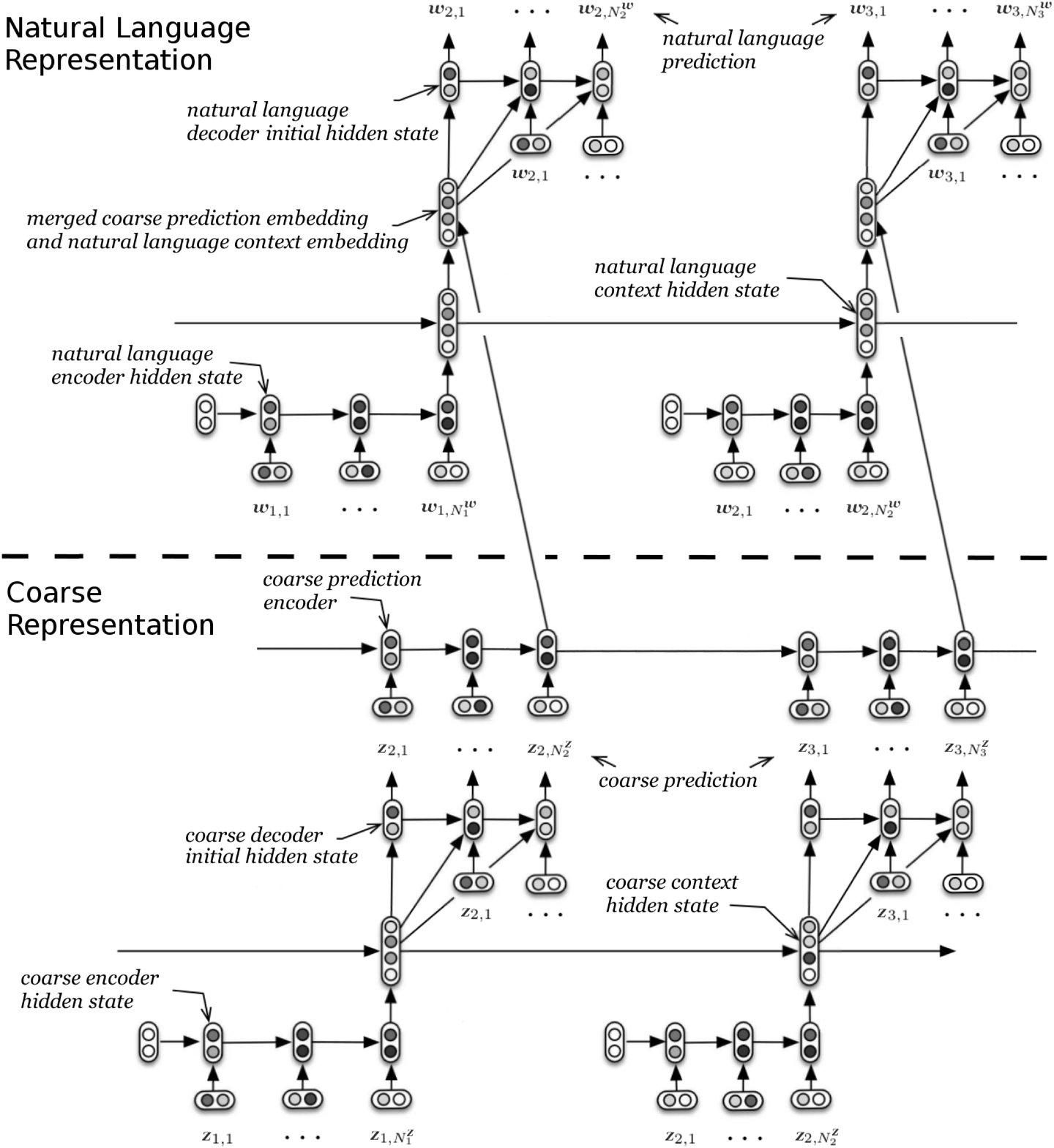}
  \caption{Computational graph for the multiresolution recurrent neural network (MrRNN). The lower part models the stochastic process over coarse tokens, and the upper part models the stochastic process over natural language tokens. The rounded boxes represent (deterministic) real-valued vectors, and the variables $z$ and $w$ represent the coarse tokens and natural language tokens respectively.}
  \label{fig:MultiresolutionHRED}
\end{figure}

We refer to the distribution over $\mathbf{z}_1, \dots, \mathbf{z}_N$ as the coarse sub-model, and to the distribution over $\mathbf{w}_1, \dots, \mathbf{w}_N$ as the natural language sub-model.
For the coarse sub-model, we parametrize the conditional distribution $P_{\theta}(\mathbf{z}_n | \mathbf{z}_1, \dots, \mathbf{z}_{n-1})$ as the HRED model described in subsection \ref{subseq:hred}, applied to the sequences $\mathbf{z}_1, \dots, \mathbf{z}_{N}$.
For the natural language sub-model, we parametrize $P_{\theta}(\mathbf{w}_n | \mathbf{w}_1, \dots, \mathbf{w}_{n-1}, \mathbf{z}_1, \dots, \mathbf{z}_{n})$ as the HRED model applied to the sequences $\mathbf{w}_1, \dots, \mathbf{w}_N$, but with one difference. The \textit{coarse prediction encoder} GRU-gated RNN encodes all the previously generated tokens $\mathbf{z}_1, \dots, \mathbf{z}_{n}$ into a real-valued vector, which is concatenated with the \textit{context} RNN and given as input to the natural language \textit{decoder} RNN.
The \textit{coarse prediction encoder} RNN is important because it encodes the high-level information, which is transmitted to the natural language sub-model.
Unlike the \textit{encoder} for the coarse-level sub-model, 
this encoding will be used to generate natural language and therefore the RNN uses different word embedding parameters.
At generation time, the coarse sub-model generates a coarse sequence (e.g.\@ a sequence of nouns), which corresponds to a high-level decision about what the natural language sequence should contain (e.g.\@ nouns to include in the natural language sequence).
Conditioned on the coarse sequence, through the \textit{coarse prediction encoder} RNN, 
the natural language sub-model then generates a natural language sequence (e.g.\@ dialogue utterance).
The model is illustrated in Figure \ref{fig:MultiresolutionHRED}.


We will assume that both $\mathbf{z}_1, \dots, \mathbf{z}_N$ and $\mathbf{w}_1, \dots, \mathbf{w}_N$ are observed and optimize the parameters w.r.t.\@ the joint log-likelihood over both sequences. At test time, to generate a response for sequence $n$ we exploit the probabilistic factorization to approximate the maximum a posteriori (MAP) estimate:
\begin{align}
& \argmax_{\mathbf{w}_n, \mathbf{z}_n} \  P_{\theta}(\mathbf{w}_n, \mathbf{z}_n | \mathbf{w}_1, \dots, \mathbf{w}_{n-1}, \mathbf{z}_1, \dots,  \mathbf{z}_{n-1}) \nonumber \\ 
& \approx \argmax_{\mathbf{w}_n} P_{\theta}(\mathbf{w}_n | \mathbf{w}_1, \dots, \mathbf{w}_{n-1}, \mathbf{z}_1, \dots,  \mathbf{z}_{n-1},  \mathbf{z}_n) \argmax_{\mathbf{z}_n} P_{\theta}(\mathbf{z}_n | \mathbf{z}_1, \dots,  \mathbf{z}_{n-1}),
\end{align}
where we further approximate the MAP for each constituent sequence using beam search. 

%% file: 40_tasks.tex
We consider the task of natural language response generation for dialogue.
Dialogue systems have been developed for applications ranging from technical support to language learning and entertainment \cite{young2013pomdp,shawar2007chatbots}. 
Dialogue systems can be categorized into two different types: goal-driven dialogue systems
and non-goal-driven dialogue systems 
\cite{DBLP:journals/corr/SerbanLCP15}.
To demonstrate the versatility of the MrRNN, we apply it to both goal-driven and non-goal-driven dialogue tasks.
We focus on the task of conditional response generation.
Given a dialogue context consisting of one or more utterances, the model must generate the next response in the dialogue.


\paragraph{Ubuntu Dialogue Corpus}

The goal-driven dialogue task we consider is technical support for the Ubuntu operating system, where we use the Ubuntu Dialogue Corpus~\cite{lowe2015ubuntu}.
The corpus consists of about $0.5$ million natural language dialogues extracted from the \textit{\#Ubuntu} Internet Relayed Chat (IRC) channel.
Users entering the chat channel usually have a specific technical problem. The users first describe their problem and afterwards other users try to help them resolve it.
The technical problems range from software-related issues (e.g.\@ installing or upgrading existing software) and hardware-related issues (e.g.\@ fixing broken drivers or partitioning hard drives) to informational needs (e.g.\@ finding software with specific functionality).
Additional details are given in appendix \ref{app:task_details}.

\paragraph{Twitter Dialogue Corpus}

The next task we consider is the non-goal-driven task of generating responses to Twitter conversations.
We use a Twitter dialogue corpus extracted in the first half of $2011$ using a procedure similar to Ritter et al.~\cite{ritter2011data}.
Unlike the Ubuntu domain,
Twitter conversations are often more noisy and do not necessarily center around a single topic.
We perform a minimal preprocessing on the dataset to remove irregular punctuation marks and afterwards tokenize it.
The dataset is split into training, validation and test sets
containing respectively $749,060$, $93,633$ and $10,000$ dialogues.\footnote{Due to Twitter's terms of service we are not allowed to redistribute Twitter content. Therefore, only the tweet IDs can be made public. These are available at: \url{www.iulianserban.com/Files/TwitterDialogueCorpus.zip}.}


%% file: 50_coarse_representations.tex
We experiment with two procedures for extracting the coarse sequence representations:

\begin{description}
\item [Noun Representation] This procedure aims to exploit the basic high-level structure of natural language discourse.
It is based on the hypothesis that dialogues are topic-driven and that these topics may be characterized by nouns.
In addition to a tokenizer, used by both the HRED and RNNLM model, it requires a part-of-speech (POS) tagger to identify the nouns in the dialogue.
The procedure uses a set of $84$ and $795$ predefined stop words for Ubuntu and Twitter respectively.
It maps a natural language utterance to its coarse representation by extracting all the nouns using the POS tagger and then removing all stop words and repeated words (keeping only the first occurrence of a word). Dialogue utterances without nouns are assigned the "no\_nouns" token. The procedure also extracts the tense of each utterance and adds it to the beginning of the coarse representation.
\item [Activity-Entity Representation] This procedure is specific to the Ubuntu technical support task, for which it aims to exploit domain knowledge related to technical problem solving. It is motivated by the observation that most dialogues are centered around \textit{activities} and \textit{entities}. For example, it is very common for users to state a specific problem they want to resolve, e.g.\@ \textit{how do I install program X?} or \textit{My driver X doesn't work, how do I fix it?}
In response to such questions, other users often respond with specific instructions, e.g.\@ \textit{Go to website X to download software Y} or \textit{Try to execute command X}.
In such cases, it is clear that the principal information resides in the technical entities and in the verbs (e.g.\@ \textit{install}, \textit{fix}, \textit{download}), and therefore that it will be advantageous to explicitly model this structure.
Motivated by this observation, the procedure uses a set of $192$ activities (verbs), created by manual inspection, and a set of $3115$ technical entities and $230$ frequent terminal commands, extracted automatically from available package managers and from the web.
The procedure uses the POS tagger to extract the verbs from the each natural language utterance.
It maps the natural language to its coarse representation by keeping only verbs from the activity set, as well as entities from the technical entity set (irrespective of their POS tags).
If no activity is found in an utterance, the representation is assigned the "none\_activity" token.
The procedure also appends a binary variable to the end of the coarse representation indicating if a terminal command was detected in the utterance.
Finally, the procedure extracts the tense of each utterance and adds it to the beginning of the coarse representation.
\end{description}

Both extraction procedures are applied at the utterance level, therefore there exists a one-to-one alignment between coarse sequences and natural language sequences (utterances).
There also exists a one-to-many alignment between the coarse sequence tokens and the corresponding natural language tokens, with the exception of a few special tokens.
Further details are given in appendix \ref{app:coarse_seq_rep}. \footnote{The pre-processed Ubuntu Dialogue Corpus used, as well as the noun representations and activity-entity representations, are available at \url{www.iulianserban.com/Files/UbuntuDialogueCorpus.zip}.}

%% file: 60_experiments.tex
The models are implemented in Theano~\cite{2016arXiv160502688short}.
We optimize all models based on the training set joint log-likelihood over coarse sequences and natural language sequences 
using the first-order stochastic gradient optimization method Adam \cite{kingma2014adampublished}. 
We train all models using early stopping with patience on the joint-log-likelihood~\cite{bengio2012practical}.
We choose our hyperparameters based on the joint log-likelihood of the validation set.
We define the $20K$ most frequent words as the vocabulary and the word embedding dimensionality to size $300$ for all models, with the exception of the RNNLM and HRED on Twitter, where we use embedding dimensionality of size $400$.
We apply gradient clipping to stop the parameters from exploding \cite{pascanu2012difficulty}.
At test time, we use a beam search of size $5$ for generating the model responses.
Further details are given in appendix \ref{app:model_details}

\subsection{Baseline Models}
We compare our models to several baselines used previously in the literature.
The first is the standard RNNLM with LSTM gating function~\cite{mikolov2010recurrent} (LSTM),
which at test time is similar to the Seq2Seq LSTM model~\cite{sutskever2014sequence}.
The second baseline is the HRED model with LSTM gating function for the \textit{decoder} RNN and GRU gating function for the \textit{encoder} RNN and \textit{context} RNN,
proposed for dialogue response generation by Serban et al.\@\cite{DBLP:conf/aaai/SerbanSBCP16} \cite{sordoni2015ahier}.
Source code for both baseline models will be made publicly available upon acceptance for publication.
For both Ubuntu and Twitter, we specify the RNNLM model to have $2000$ hidden units with the LSTM gating function.
For Ubuntu, we specify the HRED model to have $500$, $1000$ and $500$ hidden units respectively for the \textit{encoder} RNN, \textit{context} RNN and \textit{decoder} RNN.
For Twitter, we specify the HRED model to have $2000$, $1000$ and $1000$ hidden units respectively for the \textit{encoder} RNN, \textit{context} RNN and \textit{decoder} RNN.
The third baseline is the latent variable latent variable hierarchical recurrent encoder-decoder (VHRED) proposed by Serban et al.~\cite{serban2016hierarchical}.
We use the exact same VHRED models as Serban et al.~\cite{serban2016hierarchical}.

For Ubuntu, we introduce a fourth baseline, called \textit{HRED + Activity-Entity Features}, which has access to the past activity-entity pairs.
This model is similar to to the natural language sub-model of the MrRNN model, with the difference that the natural language \textit{decoder} RNN is conditioned on a real-valued vector, produced by a GRU RNN encoding \textit{only} the past coarse-level activity-entity sub-sequences.
This baseline helps differentiate between a model which observes the coarse-level sequences only as as additional features and a model which explicitly models the stochastic process of the coarse-level sequences. 
We specify the model to have $500$, $1000$, $2000$ hidden units respectively for the \textit{encoder} RNN, \textit{context} RNN and \textit{decoder} RNN. We specify the GRU RNN encoding the past coarse-level activity-entity sub-sequences to have $500$ hidden units.

\subsection{Multiresolution RNN}
The coarse sub-model is parametrized as the Bidirectional-HRED model~\cite{DBLP:conf/aaai/SerbanSBCP16} with $1000$, $1000$ and $2000$ hidden units respectively for the coarse-level \textit{encoder}, \textit{context} and \textit{decoder} RNNs.
The natural language sub-model is parametrized as a conditional HRED model with $500$, $1000$ and $2000$ hidden units respectively for the natural language \textit{encoder}, \textit{context} and \textit{decoder} RNNs. 
The \textit{coarse prediction encoder} RNN GRU RNN is parametrized with $500$ hidden units.

\subsection{Ubuntu}

\paragraph{Evaluation Methods}
It has long been known that accurate evaluation of dialogue system responses is difficult~\cite{schatzmann2005quantitative}.
Liu et al.~\cite{liu2016not} have recently shown that all automatic evaluation metrics adapted for such evaluation, including word overlap-based metrics such as BLEU and METEOR, have either very low or no correlation with human judgment of the system performance.
We therefore carry out an in-lab human study to evaluate the Ubuntu models.
We recruit $5$ human evaluators, and show them each $30-40$ dialogue contexts with the ground truth response and $4$ candidate responses (HRED, HRED + Activity-Entity Features and MrRNNs).
For each context example, we ask them to compare the candidate responses to the ground truth response and dialogue context,
and rate them for fluency and relevancy on a scale $0-4$.
Our setup is very similar to the evaluation setup used by Koehn and Monz~\cite{koehn2006manual}, and comparable to Liu et al~\cite{liu2016not}.
Further details are given in appendix \ref{app:human_eval}.

We further propose a new set of metrics for evaluating model responses on Ubuntu, which compare the activities and entities in the model generated response with those of the ground truth response.
That is, the ground truth and model responses are mapped to their respective activity-entity representations, using the automatic procedure discussed in section \ref{seq:coarse_seq_rep},
and then the overlap between their activities and entities are measured according to precision, recall and F1-score.
Based on a careful manual inspection of the extracted activities and entities, we believe that these metrics are particularly suited for the goal-oriented Ubuntu Dialogue Corpus. 
The activities and entities reflect the principal instructions given in the responses, which are key to resolving the technical problems.
Therefore, a model able to generate responses with actions and entities similar to the ground truth human responses -- which often do lead to solving the users problem -- is more likely to yield a successful dialogue system.
The reader is encouraged to verify the details and completeness of the activity-entity representations in appendix \ref{app:coarse_seq_rep}.
Scripts to generate the noun and activity-entity representations, and to evaluate the dialogue responses w.r.t.\@ activity-entity pairs are available online.\footnote{\url{https://github.com/julianser/Ubuntu-Multiresolution-Tools/tree/master/ActEntRepresentation}.}

\paragraph{Results}

\begin{table}[t]
  \caption{Ubuntu evaluation using precision (P), recall (R), F1 and accuracy metrics w.r.t.\@ activity, entity, tense and command (Cmd) on ground truth utterances, and human fluency and relevancy scores given on a scale 0-4 {\small ($^*$ indicates scores significantly different from baseline models at $90\%$ confidence)}} \label{tabel:ubuntu_results}
  \small
  \centering
    \begin{tabular}{lccccccccccccc}
    \toprule
     & \multicolumn{3}{c}{\textbf{Activity}} & \multicolumn{3}{c}{\textbf{Entity}} & \textbf{Tense} & \textbf{Cmd} & \multicolumn{2}{c}{\textbf{Human Eval.\@}} \\ \midrule
    \textbf{Model} & \textbf{P\@} & \textbf{R\@} & \textbf{F1} & \textbf{P\@} & \textbf{R\@} & \textbf{F1} & \textbf{Acc.\@} & \textbf{Acc.\@} & \textbf{Fluency} & \textbf{Relevancy}\\
    \midrule
        \parbox[c][2.65em][c]{0.085\textwidth}{LSTM} & $1.7$ & $1.03$ & $1.18$ & $1.18$ & $0.81$ & $0.87$ & $14.57$ & $94.79$ & - & - \\
        \parbox[c][2.65em][c]{0.085\textwidth}{HRED} & $5.93$ & $4.05$ & $4.34$ & $2.81$ & $2.16$ & $2.22$ & $22.2$ & $92.58$ & 2.98 & 1.01\\
        \parbox[c][2.65em][c]{0.085\textwidth}{VHRED} & $6.43$ & $4.31$ & $4.63$ & $3.28$ & $2.41$ & $2.53$ & $20.2$ & $92.02$ & - & - \\       
        \parbox[c][2.65em][c]{0.085\textwidth}{HRED + \\ Act.\@-Ent.\@} & $7.15$ & $5.5$ & $5.46$ & $3.03$ & $2.43$ & $2.44$ & $28.02$ & $86.69$ & 2.96 & 0.75 \\
        \parbox[c][2.65em][c]{0.085\textwidth}{MrRNN \\ Noun} & $5.81$ & $3.56$ & $4.04$ & $\mathbf{8.68}$ & $\mathbf{5.55}$ & $\mathbf{6.31}$ & $24.03$ & $90.66$ & $\mathbf{3.48}^*$ & $\mathbf{1.32}^*$ \\
        \parbox[c][2.65em][c]{0.085\textwidth}{MrRNN \\ Act.\@-Ent.\@} & $\mathbf{16.84}$ & $\mathbf{9.72}$ & $\mathbf{11.43}$ & $4.91$ & $3.36$ & $3.72$ & $\mathbf{29.01}$ & $\mathbf{95.04}$ & $\mathbf{3.42}^*$ & $1.04$ \\ \bottomrule
    \end{tabular}
\end{table}

The results on Ubuntu are given in table \ref{tabel:ubuntu_results}.
The MrRNNs clearly perform substantially better than the baseline models both w.r.t.\@ human evaluation and automatic evaluation metrics.
The MrRNN with noun representations achieves $2x-3x$ higher scores w.r.t.\@ entities compared to other models,
and the human evaluators also rate its fluency and relevancy substantially higher than other models.
The MrRNN with activity representations achieves $2x-3x$ higher scores w.r.t.\@ activities compared to other models and nearly $2x$ higher scores w.r.t.\@ entities compared to all baselines.
Human evaluators also rate its fluency substantially higher than the baseline models. 
However,its relevancy is rated only slightly higher compared to the HRED model, which we believe is caused by human evaluators being more likely to noticing software entities than actions in the dialogue responses (even though actions are critical to solving the actual technical problem).
Overall, the results demonstrate that the MrRNNs have learned to model high-level goal-oriented sequential structure on Ubuntu.

\begin{table}[ht]
 \caption{Ubuntu model examples. The arrows indicate change of turn.}
 \label{table:ubuntu-examples-small}
 \scriptsize
 \centering
 \begin{tabular}{p{75mm}|p{50mm}}
 \textbf{Context} & \textbf{Response} \\ \hline
         Hey guys what do you in general use for irc something ubuntu xchat or xchat-gnome ? $\rightarrow$ without -gnome. that is just cut down $\rightarrow$ you mean drop xchat-gnome and go with xchat ? & \textbf{MrRNN Act.\@ -Ent.\@:} im using xchat right now \newline \textbf{MrRNN Noun:} what is xchat-gnome ? \newline \textbf{VHRED:} correct \newline \textbf{HRED:} yes \\ \hline 
        when setting up rules with iptables command only writes changes this file " \/etc\/iptables. rules "? i ask this so i can backup before messing anything $\rightarrow$ sudo iptables-save something . dat to backup your rules restore with sudo iptables-restore \textless \ something . dat & \textbf{MrRNN Act.\@ -Ent.\@:} I don't see any reason why you need iptables to do that \newline \textbf{MrRNN Noun:} are you using ubuntu ? \newline \textbf{VHRED:} thx \newline \textbf{HRED:} thanks \\ \hline 
        Hello I tried to install ubuntu studio but I get W : Failed to fetch \textless url \textgreater 404 Not Found when apt-get update $\rightarrow$ the ppa you added is not online & \textbf{MrRNN Act.\@ -Ent.\@:} try a different mirror \newline \textbf{MrRNN Noun:} something \newline \textbf{VHRED:} but I dont know how to add it \newline \textbf{HRED:} how do I do that ? \\ \hline 
        hi community.. is there any difference between " ~/.bashrc " " ~/.bashrc. local " $\rightarrow$ yes, they're different files :-) the default ~/.bashrc sources ~/.bashrc. local $\rightarrow$ sorry I could undersatnd... If I write on terminal " gedit ~/.bashrc " it opens a file but how I open " ~/.bashrc. local " $\rightarrow$ " gedit ~/.bashrc. local "... $\rightarrow$ but it just open blank file nothing inside it & \textbf{MrRNN Act.\@ -Ent.\@:} open up gedit ~/.bashrc, there is a file called " something " \newline \textbf{MrRNN Noun:} the file is empty.. \newline \textbf{VHRED:} it's in ~/.bashrc or ~/.bashrc  \newline \textbf{HRED:} what are you trying to do ? \\
 \bottomrule
 \end{tabular}
\end{table}

Model responses are shown in Table \ref{table:ubuntu-examples-small}.
In general, the MrRNN responses are more coherent and topic-oriented compared to the other model responses, which usually produce very generic responses~\cite{DBLP:conf/aaai/SerbanSBCP16}.
In particular, the MrRNN with activity-entity representation appears to give more goal-oriented instructions compared to the MrRNN with noun representation (see examples 2-4 in Table \ref{table:ubuntu-examples-small}).
Additional examples are shown in appendix \ref{appendix:ubuntu_model_examples}.

\subsection{Twitter}

\paragraph{Evaluation Methods} 
For Twitter, similar to the Ubuntu metrics, we use the precision, recall and F1 metrics between the model responses and ground truth responses w.r.t.\@ the noun representation.
The reason we propose to use these metrics is similar to the reason given for the Ubuntu metrics related to entities: a good model response is one which includes the same nouns as the ground truth response.
We also compute the tense accuracy, as we did for Ubuntu.
Furthermore, we use the three embedding-based textual similarity metrics proposed by Liu et al.~\cite{liu2016not}:
\textit{Embedding Average} (Average), \textit{Embedding Extrema} (Extrema) and  \textit{Embedding Greedy} (Greedy).
All three metrics are based on computing the textual similarity between the ground truth response and the model response using word embeddings.
All three metrics measure topic similarity: if a model-generated response is on the same topic as the ground truth response (e.g.\@ contain paraphrases of the same words), the metrics will yield a high score.
This is a highly desirable property for dialogue systems on an open platform such as Twitter, however it is also substantially different from measuring the overall dialogue system performance, or the appropriateness of a single response, which would require human evaluation.

\begin{table}[t]
  \caption{Twitter evaluation using precision (P), recall (R), F1 and accuracy metrics w.r.t.\@ noun representation, tense accuracy and embedding-based evaluation metrics on ground truth utterances.} \label{table:TwitterResults}
  \small
  \centering
    \begin{tabular}{lccccccc}
    \toprule
     & \multicolumn{3}{c}{\textbf{Noun}} & \textbf{Tense} & \multicolumn{3}{c}{\textbf{Embedding Metrics}} \\ \midrule
    \textbf{Model} & \textbf{P\@} & \textbf{R\@} & \textbf{F1} & \textbf{Acc.\@} & \textbf{Average} & \textbf{Greedy} & \textbf{Extrema} \\
    \midrule
        \parbox[c][2.65em][c]{0.085\textwidth}{LSTM} & $0.71$ & $0.71$ & $0.65$ & $27.06$ & $51.24$ & $38.9$ & $36.58$ \\
        \parbox[c][2.65em][c]{0.085\textwidth}{HRED} & $0.31$ & $0.31$ & $0.29$ & $26.47$ & $50.1$ & $37.83$ & $35.55$ \\
        \parbox[c][2.65em][c]{0.085\textwidth}{VHRED} & $0.5$ & $0.51$ & $0.46$ & $26.66$ & $\mathbf{53.26}$ & $39.64$ & $\mathbf{37.98}$ \\       
        \parbox[c][2.65em][c]{0.085\textwidth}{MrRNN \\ Noun} & $\mathbf{4.82}$ & $\mathbf{5.22}$ & $\mathbf{4.63}$ & $\mathbf{34.48}$ & $49.77$ & $\mathbf{40.44}$ & $37.45$ \\ \bottomrule
    \end{tabular}
\end{table}


\paragraph{Results} 

The results on Twitter are given in Table \ref{table:TwitterResults}.
The responses of the MrRNN with noun representation are better than all other models on precision, recall and F1 w.r.t nouns.
MrRNN is also better than all other models w.r.t.\@ tense accuracy, and it is on par with VHRED on the embedding-based metrics.
In accordance with our previous results, this indicates that the model has learned to generate more on-topic responses and, thus, that explicitly modeling the stochastic process over nouns helps learn the high-level structure.
This is confirmed by qualitative inspection of the generated responses,
which are clearly more topic-oriented. 
See Table \ref{table:twitter-examples} in appendix.

%% file: 20_related_work.tex
Closely related to our work is the model proposed by Ji et al.\cite{ji2016latent},
which jointly models natural language text and high-level discourse phenomena.
However, it only models a discrete class per sentence at the high level, which must be manually annotated by humans.
On the other hand, MrRNN models a sequence of automatically extracted high-level tokens.
Recurrent neural network models with stochastic latent variables, such as the Variational Recurrent Neural Networks by Chung et al.~\cite{chung2015recurrent}, are also closely related to our work.
These models face the more difficult task of learning the high-level representations,
while simultaneously learning to model the generative process over high-level sequences and low-level sequences,
which is a more difficult optimization problem.
In addition to this, such models assume the high-level latent variables to be continuous, usually Gaussian, distributions.

Recent dialogue-specific neural network architectures, such as the model proposed by Wen et al.~\cite{wen2016network}, are also relevant to our work.
Different from the MrRNN, they require domain-specific hand-crafted high-level (dialogue state) representations with human-labelled examples, and they usually consist of several sub-components each trained with a different objective function.


%% file: 70_discussion.tex
We have proposed the multiresolution recurrent neural network (MrRNN) for generatively modeling sequential data at multiple levels of abstraction. It is trained by optimizing the joint log-likelihood over the sequences at each level.
We apply MrRNN to dialog response generation on two different tasks, Ubuntu technical support and Twitter conversations, and evaluate it in a human evaluation study and via automatic evaluation metrics.
On Ubuntu, MrRNN demonstrates dramatic improvements compared to competing models.
On Twitter, MrRNN appears to generate more relevant and on-topic responses.
Even though abstract information is implicitly present in natural language dialogues,
by explicitly representing information at different levels of abstraction and jointly optimizing the generation process across abstraction levels,
MrRNN is able to generate more fluent, relevant and goal-oriented responses.
The results suggest that the fine-grained abstraction (low-level) provides the architecture with increased fluency for predicting natural utterances,
while the coarse-grained (high-level) abstraction gives it the semantic structure necessary to generate more coherent and relevant utterances.
The results also imply that it is not simply a matter of adding additional features for prediction -- MrRNN outperforms a competitive baseline augmented with the coarse-grained abstraction sequences as features -- rather, it is the combination of representation and generation at multiple levels that yields the improvements.
Finally, we observe that the architecture provides a general framework for modeling discrete sequences, as long as a coarse abstraction is available.
We therefore conjecture that the architecture may successfully be 
applied to broader natural language generation tasks, such as generating prose and persuasive argumentation, and other tasks involving discrete sequences, such as music composition.
We leave this to future work.

%% file: appendix.tex
\section{Task Details} \label{app:task_details}

\paragraph{Ubuntu} We use the Ubuntu Dialogue Corpus v2.0 extracted Jamuary, 2016: \url{http://cs.mcgill.ca/~jpineau/datasets/ubuntu-corpus-1.0/}.

\paragraph{Twitter} We preprocess the dataset using the Moses tokenizer extracted June, 2015: \url{https://github.com/moses-smt/mosesdecoder/blob/master/scripts/tokenizer/tokenizer.perl}.\footnote{Due to Twitter's Terms and Conditions we are unfortunately not allowed to publish the preprocessed dataset.}

\section{Coarse Sequence Representations} \label{app:coarse_seq_rep}







\subsection*{Nouns}

The noun-based procedure for extracting coarse tokens aims to exploit high-level structure of natural language discourse. More specifically, it builds on the hypothesis that dialogues in general are topic-driven and that these topics may be characterized by the nouns inside the dialogues. At any point in time, the dialogue is centered around one or several topics. As the dialogue progresses, the underlying topic evolves as well. In addition to the tokenizer required by the previous extraction procedure, this procedure also requires a part-of-speech (POS) tagger to identify the nouns in the dialogue suitable for the language domain.

For extracting the noun-based coarse tokens, we define a set of $795$ stop words for Twitter and $84$ stop words for Ubuntu containing mainly English pronouns, punctuation marks and prepositions (excluding special placeholder tokens). We then extract the coarse tokens by applying the following procedure to each dialogue:

\begin{enumerate}
\item We apply the POS tagger version $0.3.2$ developed by Owoputi and colleagues~\cite{owoputi2013improved} to extract POS.\footnote{\url{www.cs.cmu.edu/~ark/TweetNLP/}} 
For Twitter, we use the parser trained on the Twitter corpus developed by Ritter et al.~\cite{Ritter:2011:NER:2145432.2145595}. 
For Ubuntu, we use the parser trained on the NPS Chat Corpus developed by Forsyth and Martell
which was extracted from IRC chat channels similar to the Ubuntu Dialogue Corpus.\footnote{As input to the POS tagger, we replace all unknown tokens with the word "something" and remove all special placeholder tokens (since the POS tagger was trained on a corpus without these words). We further reduce any consecutive sequence of spaces to a single space. For Ubuntu, we also replace all commands and entities with the word "something". For Twitter, we also replace all numbers with the word "some", all urls with the word "somewhere" and all heart \textit{emoticons} with the word "love".}\footnote{Forsyth, E. N. and Martell, C. H. (2007). Lexical and discourse analysis of online chat dialog. In Semantic
Computing, 2007. ICSC 2007. International Conference on, pages 19–26. IEEE.}
\item Given the POS tags, we remove all words which are not tagged as nouns and all words containing non-alphabet characters.\footnote{We define nouns as all words with tags containing the prefix "NN" according to the PTB-style tagset.}. We keep all urls and paths.
\item We remove all stop words and all repeated tokens, while maintaining the order of the tokens.
\item We add the "no\_nouns" token to all utterances, which do not contain any nouns. This ensures that no coarse sequences are empty. It also forces the coarse sub-model to explicitly generate at least one token, even when there are no actual nouns to generate.
\item For each utterance, we use the POS tags to detect three types of time tenses: past, present and future tenses. We append a token indicating which of the $2^3$ tenses are present at the beginning of each utterance.\footnote{Note that an utterance may contain several sentences. It therefore often happens that an utterance contains several time tenses.} If no tenses are detected, we append the token "no\_tenses".
\end{enumerate}

As before, there exists a one-to-many alignment between the extracted coarse sequence tokens and the natural language tokens, since this procedure also maintains the ordering of all special placeholder tokens, with the exception of the "no\_nouns" token.

We cut-off the vocabulary at $10000$ coarse tokens for both the Twitter and Ubuntu datasets excluding the special placeholder tokens. On average a Twitter dialogue in the training set contains $~25$ coarse tokens, while a Ubuntu dialogue in the training set contains $~36$ coarse tokens.

Model statistics for the unigram and bigram language models are presented in Table \ref{table:noun-rep-bits-per-word} for the noun representations on the Ubuntu and Twitter training sets.\footnote{The models were trained using maximum log-likelihood on the noun representations excluding all special tokens.}
The table shows a substantial difference in bits per words between the unigram and bigram models, which suggests that the nouns are significantly correlated with each other.

\begin{table}[t]
  \caption{Unigram and bigram models bits per word on noun representations.}
  \label{table:noun-rep-bits-per-word}
  \small
  \centering
  \begin{tabular}{lcc}
    \toprule
    Model & \textbf{Ubuntu} & \textbf{Twitter} \\ 
    \midrule
    Unigram & 10.16 & 12.38 \\
    Bigram & 7.26 & 7.76 \\
    \bottomrule
  \end{tabular}
 \end{table}
 
\subsection*{Activity-Entity Pairs}

The activity-entity-based procedure for extracting coarse tokens attempts to exploit domain specific knowledge for the Ubuntu Dialogue Corpus, in particular in relation to providing technical assistance with problem solving. Our manual inspection of the corpus shows that many dialogues are centered around \textit{activities}. For example, it is very common for users to state a specific problem they want to resolve, e.g.\@ \textit{how do I install program X?} or \textit{My driver X doesn't work, how do I fix it?}. In response to such queries, other users often respond with specific instructions, e.g.\@ \textit{Go to website X to download software Y} or \textit{Try to execute command X}. In addition to the technical entities, the principle message conveyed by each utterance resides in the verbs, e.g.\@ \textit{install}, \textit{work}, \textit{fix}, \textit{go}, \textit{to}, \textit{download}, \textit{execute}. Therefore, it seems clear that a dialogue system must have a strong understanding of both the activities and technical entities if it is to effectively assist users with technical problem solving. It seems likely that this would require a dialogue system able to relate technical entities to each other, e.g.\@ to understand that \textit{firefox} depends on the \textit{GCC} library, and conform to the temporal structure of activities, e.g.\@ understanding that the \textit{install} activity is often followed by \textit{download} activity.

We therefore construct two word lists: one for activities and one for technical entities. We construct the activity list based on manual inspection yielding a list of $192$ verbs. For each activity, we further develop a list of synonyms and conjugations of the tenses of all words. We also use Word2Vec word embeddings \cite{mikolov2013distributedbetter}, trained on the Ubuntu Dialogue Corpous training set, to identify commonly misspelled variants of each activity. The result is a dictionary, which maps a verb to its corresponding activity (if such exists). For constructing the technical entity list, we scrape publicly available resources, including Ubuntu and Linux-related websites as well as the Debian package manager \textit{APT}. Similar to the activities, we also use the Word2Vec word embeddings to identify misspelled and paraphrased entities. This results in another dictionary, which maps one or two words to the corresponding technical entity. In total there are $3115$ technical entities. In addition to this we also compile a list of $230$ frequent commands. Examples of the extracted activities, entities and commands can be found in the appendix.

Afterwards, we extract the coarse tokens by applying the following procedure to each dialogue:

\begin{enumerate}
\item We apply the technical entity dictionary to extract all technical entities.
\item We apply the POS tagger version $0.3.2$ developed by Owoputi and colleagues, trained on the NPS Chat Corpus developed by Forsyth and Martell as before. As input to the POS tagger, we map all technical entities to the token "something". This transformation should improve the POS tagging accuracy, since The corpus the parser was trained on does not contain technical words.
\item Given the POS tags, we extract all verbs which correspond to activities.\footnote{We define verbs as all words with tags containing the prefix "VB" according to the PTB-style tagset.}. If there are no verbs in an entire utterance and the POS tagger identified the first word as a noun, we will assume that the first word is in fact a verb. We do this, because the parser does not work well for tagging technical instructions in imperative form, e.g.\@ \textit{upgrade firefox}. If no activities are detected, we append the token "none\_activity" to the coarse sequence. We also keep all urls and paths.
\item We remove all repeated activities and technical entities, while maintaining the order of the tokens.
\item If a command is found inside an utterance, we append the "cmd" token at the end of the utterance. Otherwise, we append the "no\_cmd" token to the end of the utterance. This enables the coarse sub-model to predict whether or not an utterance contains executable commands.
\item As for the noun-based coarse representation, we also append the time tense to the beginning of the sequence.
\end{enumerate}

As before, there exists a one-to-many alignment between the extracted coarse sequence tokens and the natural language tokens, with the exception of the "none\_activity" and "no\_cmd" tokens.

Since the number of unique tokens are smaller than $10000$, we do not need to cut-off the vocabulary. On average a Ubuntu dialogue in the training set contains $~43$ coarse tokens.

Our manual inspection of the extracted coarse sequences, show that the technical entities are identified with very high accuracy and that the activities capture the main intended action in the majority of utterances. Due to the high quality of the extracted activities and entities, we are confident that they may be used for evaluation purposes as well. 

Scripts to generate the noun and activity-entity representations, and to evaluate the dialogue responses w.r.t.\@ activity-entity pairs are available online at: \url{https://github.com/julianser/Ubuntu-Multiresolution-Tools/tree/master/ActEntRepresentation}.

\begin{table}[t]
  \caption{Twitter Coarse Sequence Examples}
  \label{TwitterCoarseSeqExamples}
  \centering
  \small
  \begin{tabular}{p{40mm}p{40mm}}
    Natural Language Tweets & Noun Representation \\
    \toprule
    \textless first\_speaker\textgreater \ at pinkberry with my pink princess enjoying a precious moment 	\textless url\textgreater \newline \newline \textless second\_speaker\textgreater - they are adorable, alma still speaks about emma bif sis . hugs  & present\_tenses pinkberry princess moment \newline \newline present\_tenses alma emma bif sis hugs    \\
    \midrule
    \textless first\_speaker\textgreater \  \textless at\textgreater \  when you are spray painting, where are you doing it ? outside ? in your apartment ? where ? \newline \newline
\textless second\_speaker\textgreater \  \textless at\textgreater \  mostly spray painting outside but
some little stuff in the bathroom .
&
present\_tenses spray painting apartment \newline \newline
present\_tenses spray stuff bathroom \\
    \bottomrule
  \end{tabular}
\end{table}

\begin{table}[t]
  \caption{Ubuntu Coarse Sequence Examples}
  \label{UbuntuCoarseSeqExamples}
  \centering
  \small
  \begin{tabular}{p{60mm}|p{60mm}}
    Natural Language Dialogues & Activity-Entity Coarse Dialogues \\
    \toprule
if you can get a hold of the logs, there 's stuff from **unknown** about his inability to install amd64
\newline \newline
I'll check fabbione 's log, thanks
sounds like he had the same problem I did
ew, why ?
...
\newline \newline
upgrade lsb-base and acpid
\newline \newline
i'm up to date
\newline \newline
what error do you get ?
\newline \newline
i don't find error :/ where do i search from ?
acpid works, but i must launch it manually in a root sterm
...
&
future\_tenses get\_activity install\_activity amd64\_entity no\_cmd
\newline \newline
no\_tenses check\_activity no\_cmd
past\_present\_tenses none\_activity no\_cmd
no\_tenses none\_activity no\_cmd
...
\newline \newline
no\_tenses upgrade\_activity lsb\_entity acpid\_entity no\_cmd
\newline \newline
no\_tenses none\_activity no\_cmd
\newline \newline
present\_tenses get\_activity no\_cmd
\newline \newline
present\_tenses discover\_activity no\_cmd
present\_future\_tenses work\_activity acpid\_entity root\_entity no\_cmd
... \\
    \bottomrule
  \end{tabular}
\end{table}

\newpage

\subsection*{Stop Words for Noun-based Coarse Tokens}

\textbf{Ubuntu stop words for noun-based coarse representation:}
\begin{framed}
\scriptsize
all another any anybody anyone anything both each each other either everybody everyone everything few he her hers herself him himself his I it its itself many me mine more most much myself neither no one nobody none nothing one one another other others ours ourselves several she some somebody someone something that their theirs them themselves these they this those us we what whatever which whichever who whoever whom whomever whose you your yours yourself yourselves . , ? ' - -- !
\end{framed}

\newpage

\textbf{Twitter stop words for noun-based coarse representation:} \footnote{Part of these were extracted from \url{https://github.com/defacto133/twitter-wordcloud-bot/blob/master/assets/stopwords-en.txt}.}
\begin{framed}
\scriptsize
all another any anybody anyone anything both each each other either everybody everyone everything few he her hers herself him himself his I it its itself many me mine more most much myself neither no one nobody none nothing one one another other others ours ourselves several she some somebody someone something that their theirs them themselves these they this those us we what whatever which whichever who whoever whom whomever whose you your yours yourself yourselves . , ? ' - -- !able about above abst accordance according accordingly across act actually added adj adopted affected affecting affects after afterwards again against ah all almost alone along already also although always am among amongst an and announce another any anybody anyhow anymore anyone anything anyway anyways anywhere apparently approximately are aren arent arise around as aside ask asking at auth available away awfully b back bc be became because become becomes becoming been before beforehand begin beginning beginnings begins behind being believe below beside besides between beyond biol bit both brief briefly but by c ca came can cannot can't cant cause causes certain certainly co com come comes contain containing contains cos could couldnt d date day did didn didn't different do does doesn doesn't doing don done don't dont down downwards due during e each ed edu effect eg eight eighty either else elsewhere end ending enough especially et et-al etc even ever every everybody everyone everything everywhere ex except f far few ff fifth first five fix followed following follows for former formerly forth found four from further furthermore g game gave get gets getting give given gives giving go goes going gone gonna good got gotten great h had happens hardly has hasn hasn't have haven haven't having he hed hence her here hereafter hereby herein heres hereupon hers herself hes hey hi hid him himself his hither home how howbeit however hundred i id ie if i'll im immediate immediately importance important in inc indeed index information instead into invention inward is isn isn't it itd it'll its itself i've j just k keep keeps kept keys kg km know known knows l ll largely last lately later latter latterly least less lest let lets like liked likely line little ll 'll lol look looking looks lot ltd m made mate mainly make makes many may maybe me mean means meantime meanwhile merely mg might million miss ml more moreover most mostly mr mrs much mug must my myself n na name namely nay nd near nearly necessarily necessary need needs neither never nevertheless new next nine ninety no nobody non none nonetheless noone nor normally nos not noted nothing now nowhere o obtain obtained obviously of off often oh ok okay old omitted omg on once one ones only onto or ord other others otherwise ought our ours ourselves out outside over overall owing own p page pages part particular particularly past people per perhaps placed please plus poorly possible possibly potentially pp predominantly present previously primarily probably promptly proud provides put q que quickly quite qv r ran rather rd re readily really recent recently ref refs regarding regardless regards related relatively research respectively resulted resulting results right rt run s said same saw say saying says sec section see seeing seem seemed seeming seems seen self selves sent seven several shall she shed she'll shes should shouldn shouldn't show showed shown showns shows significant significantly similar similarly since six slightly so some somebody somehow someone somethan something sometime sometimes somewhat somewhere soon sorry specifically specified specify specifying state states still stop strongly sub substantially successfully such sufficiently suggest sup sure t take taken taking tbh tell tends th than thank thanks thanx that that'll thats that've the their theirs them themselves then thence there thereafter thereby thered therefore therein there'll thereof therere theres thereto thereupon there've these they theyd they'll theyre they've thing things think this those thou though thoughh thousand throug through throughout thru thus til time tip to together too took toward towards tried tries truly try trying ts tweet twice two u un under unfortunately unless unlike unlikely until unto up upon ups ur us use used useful usefully usefulness uses using usually v value various ve 've very via viz vol vols vs w wanna want wants was wasn wasn't way we wed welcome well we'll went were weren weren't we've what whatever what'll whats when whence whenever where whereafter whereas whereby wherein wheres whereupon wherever whether which while whim whither who whod whoever whole who'll whom whomever whos whose why widely will willing wish with within without won won't words world would wouldn wouldn't www x y yeah yes yet you youd you'll your youre yours yourself yourselves you've z zero
\end{framed}

\subsection*{Activities and Entities for Ubuntu Dialogue Corpus} 

\textbf{Ubuntu activities:}
\begin{framed}
\scriptsize
accept, activate, add, ask, appoint, attach, backup, boot, check, choose, clean, click, comment, compare, compile, compress, change, affirm, connect, continue, administrate, copies, break, create, cut, debug, decipher, decompress, define, describe, debind, deattach, deactivate, download, adapt, eject, email, conceal, consider, execute, close, expand, expect, export, discover, correct, fold, freeze, get, deliver, go, grab, hash, import, include, install, interrupt, load, block, log, log-in, log-out, demote, build, clock, bind, more, mount, move, navigate, open, arrange, partition, paste, patch, plan, plug, post, practice, produce, pull, purge, push, put, queries, quote, look, reattach, reboot, receive, reject, release, remake, delete, name, replace, request, reset, resize, restart, retry, return, revert, reroute, scroll, send, set, display, shutdown, size, sleep, sort, split, come-up, store, signup, get-ahold-of, say, test, transfer, try, uncomment, de-expand, uninstall, unmount, unplug, unset, sign-out, update, upgrade, upload, use, delay, enter, support, prevent, loose, point, contain, access, share, buy, sell, help, work, mute, restrict, play, call, thank, burn, advice, force, repeat, stream, respond, browse, scan, restore, design, refresh, bundle, implement, programming, compute, touch, overheat, cause, affect, swap, format, rescue, zoomed, detect, dump, simulate, checkout, unblock, document, troubleshoot, convert, allocate, minimize, maximize, redirect, maintain, print, spam, throw, sync, contact, destroy
\end{framed}

\newpage

\textbf{Ubuntu entities (excerpt):}
\begin{framed}
\scriptsize
ubuntu\_7.04, dmraid, vnc4server, tasksel, aegis, mirage, system-config-audit, uif2iso, aumix, unrar, dell, hibernate, ucoded, finger, zoneminder, ucfg, macaddress, ia32-libs, synergy, aircrack-ng, pulseaudio, gnome, kid3, bittorrent, systemsettings, cups, finger, xchm, pan, uwidget, vnc-java, linux-source, ucommand.com, epiphany, avanade, onboard, uextended, substance, pmount, lilypond, proftpd, unii, jockey-common, aha, units, xrdp, mp3check, cruft, uemulator, ulivecd, amsn, ubuntu\_5.10, acpidump, uadd-on, gpac, ifenslave, pidgin, soundconverter, kdelibs-bin, esmtp, vim, travel, smartdimmer, uactionscript, scrotwm, fbdesk, tulip, beep, nikto, wine, linux-image, azureus, vim, makefile, uuid, whiptail, alex, junior-arcade, libssl-dev, update-inetd, uextended, uaiglx, sudo, dump, lockout, overlay-scrollbar, xubuntu, mdk, mdm, mdf2iso, linux-libc-dev, sms, lm-sensors, dsl, lxde, dsh, smc, sdf, install-info, xsensors, gutenprint, sensors, ubuntu\_13.04, atd, ata, fatrat, fglrx, equinix, atp, atx, libjpeg-dbg, umingw, update-inetd, firefox, devede, cd-r, tango, mixxx, uemulator, compiz, libpulse-dev, synaptic, ecryptfs, crawl, ugtk+, tree, perl, tree, ubuntu-docs, libsane, gnomeradio, ufilemaker, dyndns, libfreetype6, daemon, xsensors, vncviewer, vga, indicator-applet, nvidia-173, rsync, members, qemu, mount, rsync, macbook, gsfonts, synaptic, finger, john, cam, lpr, lpr, xsensors, lpr, lpr, screen, inotify, signatures, units, ushareware, ufraw, bonnie, nec, fstab, nano, bless, bibletime, irssi, ujump, foremost, nzbget, ssid, onboard, synaptic, branding, hostname, radio, hotwire, xebia, netcfg, xchat, irq, lazarus, pilot, ucopyleft, java-common, vm, ifplugd, ncmpcpp, irc, uclass, gnome, sram, binfmt-support, vuze, java-common, sauerbraten, adapter, login
\end{framed}

\textbf{Ubuntu commands:}
\begin{framed}
\scriptsize
alias, apt-get, aptitude, aspell, awk, basename, bc, bg, break, builtin, bzip2, cal, case, cat, cd, cfdisk, chgrp, chmod, chown, chroot, chkconfig, cksum, cmp, comm, command, continue, cp, cron, crontab, csplit, curl, cut, date, dc, dd, ddrescue, declare, df, diff, diff3, dig, dir, dircolors, dirname, dirs, dmesg, du, echo, egrep, eject, enable, env, eval, exec, exit, expect, expand, export, expr, false, fdformat, fdisk, fg, fgrep, file, find, fmt, fold, for, fsck, ftp, function, fuser, gawk, getopts, grep, groupadd, groupdel, groupmod, groups, gzip, hash, head, history, hostname, htop, iconv, id, if, ifconfig, ifdown, ifup, import, install, ip, jobs, join, kill, killall, less, let, link, ln, local, locate, logname, logout, look, lpc, lpr, lprm, ls, lsof, man, mkdir, mkfifo, mknod, more, most, mount, mtools, mtr, mv, mmv, nc, nl, nohup, notify-send, nslookup, open, op, passwd, paste, ping, pkill, popd, pr, printf, ps, pushd, pv, pwd, quota, quotacheck, quotactl, ram, rar, rcp, read, readonly, rename, return, rev, rm, rmdir, rsync, screen, scp, sdiff, sed, select, seq, set, shift, shopt, shutdown, sleep, slocate, sort, source, split, ssh, stat, strace, su, sudo, sum, suspend, sync, tail, tar, tee, test, time, timeout, times, touch, top, tput, traceroute, tr, true, tsort, tty, type, ulimit, umask, unalias, uname, unexpand, uniq, units, unrar, unset, unshar, until, useradd, userdel, usermod, users, uuencode, uudecode, vi, vmstat, wait, watch, wc, whereis, which, while, who, whoami, write, xargs, xdg-open, xz, yes, zip, admin, purge
\end{framed}

\newpage

\section{Model Details} \label{app:model_details}

\subsection*{Training}

All models were trained with a learning rate of $0.0002$ or $0.0001$, batches of size either $40$ or size $80$ and gradients are clipped at $1.0$.
We truncate the backpropagation to batches with $80$ tokens
We validate on the entire validation set every $5000$ training batches.
We choose almost identical hyperparameters for the Ubuntu and Twitter models, since the models appear to perform similarly w.r.t.\@ different hyperparameters and since the statistics of the two datasets are comparable.
We use the $20K$ most frequent words on Twitter and Ubuntu as the natural language vocabulary for all the models, and assign all words outside the vocabulary to a special unknown token symbol.
For MrRNN, we use a coarse token vocabulary consisting of the $10K$ most frequent tokens in the coarse token sequences.

\subsection*{Generation}
We compute the cost of each beam search (candidate response) as the log-likelihood of the tokens in the beam divided by the number of tokens it contains.
The LSMT model performs better when the beam search is not allowed to generate the unknown token symbol, however even then it still performs worse than the HRED model across all metrics except for the command accuracy.

\subsection*{Baselines}

Based on preliminary experiments,
we found that a slightly different parametrization of the HRED baseline model worked better on Twitter.
The \textit{encoder} RNN has a bidirectional GRU RNN encoder, with $1000$ hidden units for the forward and backward RNNs each, and a \textit{context} RNN and a \textit{decoder} RNN with $1000$ hidden units each.
Furthermore, the \textit{decoder} RNN computes a $1000$ dimensional real-valued vector for each hidden time step, which is multiplied with the output \textit{context} RNN. The output is feed through a one-layer feed-forward neural network with hyperbolic tangent activation function, which the \textit{decoder} RNN then conditions on.

\section{Human Evaluation} \label{app:human_eval}

\begin{figure}[ht]
  \centering
  \includegraphics[width=0.5\linewidth]{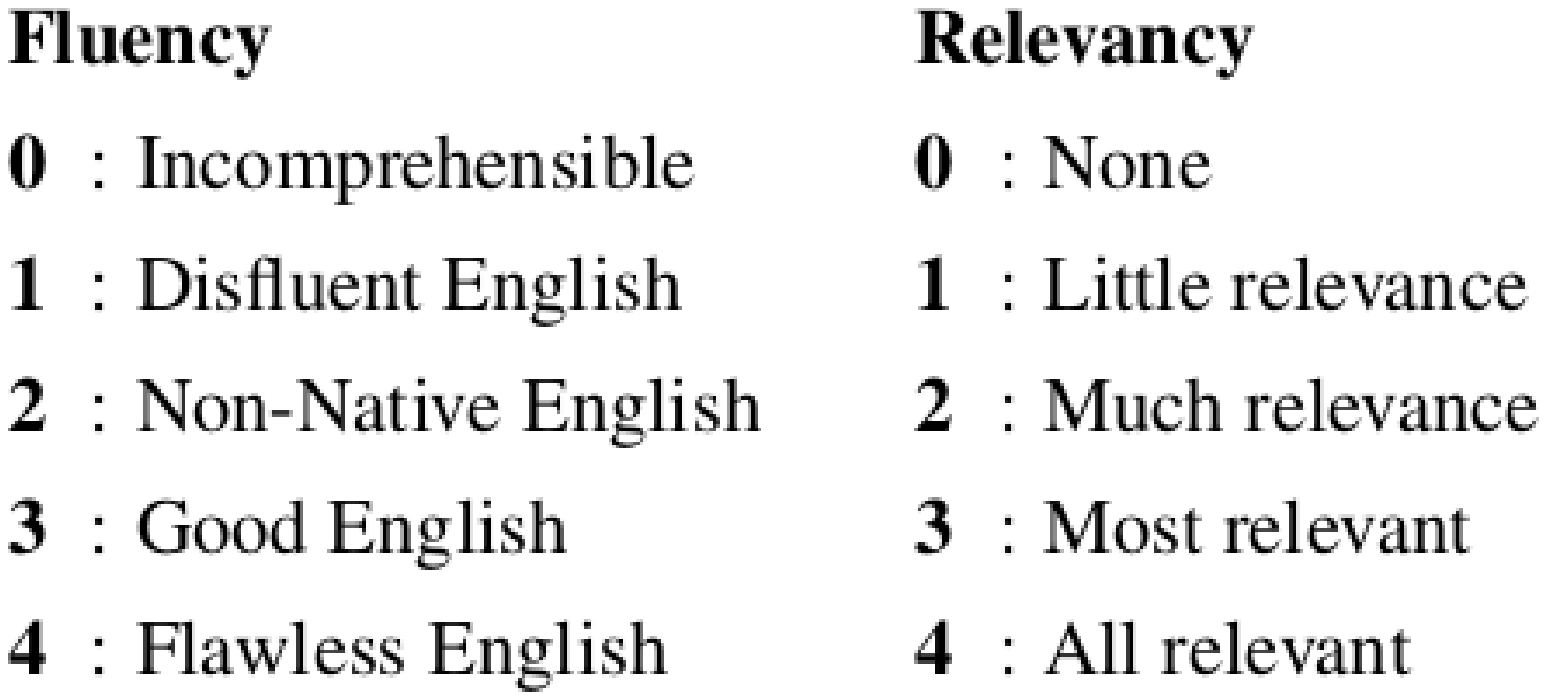}
  \caption{Fluency and relevancy reference table presented to human evaluators.}
  \label{tab:fluency_relevancy_table}
\end{figure}

All human evaluators either study or work in an English speaking environment,
and have indicated that they have some experience using a Linux operating system.
To ensure a high quality of the ground truth responses, human evaluators were only asked to evaluate responses, where the ground truth contained at least one technical entity.
Before starting evaluators, were shown one short annotated example with a brief explanation of how to give annotations.
In particular, the evaluators were instructed to use the following reference in Figure \ref{tab:fluency_relevancy_table}.

The $5$ evaluators gave $1069$ ratings in total.
Table \ref{tabel:ubuntu_human_scores_results} shows the scores by category.
\begin{table}[ht]
  \caption{Ubuntu human fluency and relevancy scores by rating category} \label{tabel:ubuntu_human_scores_results}
  \small
  \centering
    \begin{tabular}{lcccccccccccccc}
    \toprule
     & \multicolumn{5}{c}{\textbf{Fluency}} & & \multicolumn{5}{c}{\textbf{Relevancy}} \\ \midrule
        \color{white} \backslashbox{\color{black} \textbf{Model}}{\color{black} \textbf{Rating Level}} \color{black} & $0$ & $1$ & $2$ & $3$ & $4$ & & $0$ & $1$ & $2$ & $3$ & $4$ \\     \midrule
        HRED & $3$ & $11$ & $21$ & $50$ & $49$ & & $68$ & $22$ & $19$ & $19$ & $4$ \\
        HRED + Act.-Ent. & $3$ & $17$ & $19$ & $37$ & $57$ & & $69$ & $39$ & $18$ & $6$ & $2$ \\
        MrRNN Noun & $1$ & $2$ & $8$ & $52$ & $71$ & & $51$ & $45$ & $24$ & $10$ & $4$ \\
        MrRNN Act.-Ent & $0$ & $2$ & $6$ & $52$ & $74$ & & $27$ & $53$ & $39$ & $14$ & $1$ \\ \bottomrule
    \end{tabular}
\end{table}

\section{Model Examples} \label{appendix:ubuntu_model_examples}

Model response examples are given in this section.
All the model responses can be downloaded at \url{www.iulianserban.com/Files/TwitterDialogueCorpus.zip} and \url{www.iulianserban.com/Files/UbuntuDialogueCorpus.zip}.

\begin{table}[ht]
 \caption{Ubuntu model examples. The $\rightarrow$ token indicates a change of turn.}
 \label{table:ubuntu-examples-first}
 \scriptsize
 \centering
 \begin{tabular}{p{65mm}|p{60mm}}
 \textbf{Context} & \textbf{Response} \\ \hline
        seriously... 50\% cpu from xorg running just two apps(terminal and system monitor ) i've been seeing this ever since i upgraded to 12.10 anyone else seeing this $\rightarrow$ I would do such kind of tests with ' top ' or ' something Enjoy the difference... $\rightarrow$ that is top saying 50\% cpu from xorg $\rightarrow$ Ok. In that case I would try a window manager without composite for a try. 50\% is too much. & \textbf{MrRNN Act.\@ -Ent.\@:} Are you using XGL ? \newline \textbf{MrRNN Noun:} What window manager are you using ? \newline \textbf{HRED + Act.\@-Ent.\@:} I don't know how to help you with that, sorry. \newline \textbf{HRED:} i have no idea how to do that, i just want to know how to do it \newline \textbf{LSTM:} something ? \\ \hline 
        my linux seems to be stuck at rev 18... other systems in the house are as high as rev 34... any suggestions ? $\rightarrow$ what 's rev ? $\rightarrow$ the linux info says kernel linux says 3.5. something generic & \textbf{MrRNN Act.\@ -Ent.\@:} how do i compile the kernel ? \newline \textbf{MrRNN Noun:} the linux kernel is 2.6. something \newline \textbf{HRED + Act.\@-Ent.\@:} how do i find out what version of ubuntu i have ? \newline \textbf{HRED:} that 's not what you want to do \newline \textbf{LSTM:} ? what do you mean ? \\ \hline 
        i need some SERIOUS help with fullscreen video playback. i already asked in \#lubuntu, but i might have a better chance here : most video files either let the player hang or audio/video goes badly out of sync right away. i know that it does work with the same files and it even did so like two hours ago, but after some browsing and man page reading, the problem comes up again. i have NO idea where to look anymore nor what I did to ' reproduce ' the $\rightarrow$ even with vlc & \textbf{MrRNN Act.\@ -Ent.\@:} works fine in vlc \newline \textbf{MrRNN Noun:} do you use vlc ? \newline \textbf{HRED + Act.\@-Ent.\@:} what video card do you have ? \newline \textbf{HRED:} yeah, i know, but i don't know how to do that, but i don't know how to do that \newline \textbf{LSTM:} ? what do you mean ? \\ \hline 
        I'm in the process of building a new desktop and I was wondering if there was any way to, using KVM, start or restart a Windows session when my girlfriend logs in. I have plenty of memory and my i5 3570 supports all the necessary virtualisation technologies. $\rightarrow$ you can install VirtualBox, and I'm sure there 's a command line method to run VB with a specified installed VM operating system, and you can make that command a startup command ending with " \& " & \textbf{MrRNN Act.\@ -Ent.\@:} have you installed virtualbox ? \newline \textbf{MrRNN Noun:} are you trying to install ubuntu ? \newline \textbf{HRED + Act.\@-Ent.\@:} I don't know how to do that, sorry. I don't know how to do that. \newline \textbf{HRED:} thanks, I'll give it a shot. \newline \textbf{LSTM:} thanks, I'll try that. \\ \hline 
        hello all, how can i setup two default gateways graphically in network manager ? $\rightarrow$ you can't have 2 default gateway on one device...? $\rightarrow$ well.. i can $\rightarrow$ hows that gonna work ? & \textbf{MrRNN Act.\@ -Ent.\@:} what are you looking for ? \newline \textbf{MrRNN Noun:} do you have a default gateway ? \newline \textbf{HRED + Act.\@-Ent.\@:} what are you trying to do ? \newline \textbf{HRED:} i want to set up a static ip, but i dont know how to set it up \newline \textbf{LSTM:} ? \\ \hline 
        They are mounted via fstab $\rightarrow$ sure they are mounted ? $\rightarrow$ I've been using them all day :[ p :P * & \textbf{MrRNN Act.\@ -Ent.\@:} I see \newline \textbf{MrRNN Noun:} something :/ something ls -ld something \newline \textbf{HRED + Act.\@-Ent.\@:} I can mount it, but when I try to mount it, it says " mount : special device /dev/sda1 does not exist \newline \textbf{HRED:} what do you mean by mount ? \newline \textbf{LSTM:} ? what are you talking about ? \\ \hline
 \end{tabular}
\end{table}

\begin{table}[ht]
 \caption{Ubuntu model examples. The $\rightarrow$ token indicates a change of turn.}
 \label{table:ubuntu-examples-second}
 \scriptsize
 \centering
 \begin{tabular}{p{65mm}|p{60mm}}
 \textbf{Context} & \textbf{Response} \\ \hline
        can you have gnome and KDE installed and use one or the other ? cough erin brainwaves & \textbf{MrRNN Act.\@-Ent.:} what are you looking for ? \newline \textbf{MrRNN Noun:} gconf-editor \newline \textbf{HRED + Act.\@-Ent.\@:} what do you want me to do ? \newline \textbf{HRED:} yeah, i just did a fresh install of ubuntu, and it worked fine. \newline \textbf{LSTM:} ? \\ \hline 
        Ugh so my mount points need to be in /media rather than /mnt to make them appear in the nautilus side-pane :\ $\rightarrow$ you could symlink /mnt to /media afaik & \textbf{MrRNN Act.\@ -Ent.\@:} thank you :) \newline \textbf{MrRNN Noun:} you need to mount it in /mnt or /media \newline \textbf{HRED + Act.\@-Ent.\@:} if you don't know what you are trying to do, then you should be able to mount it manually, but I don't know how \newline \textbf{HRED:} I don't know how to do that, I just want to be able to mount them in /media \newline \textbf{LSTM:} what ? \\ \hline 
        How can i install ubuntu raring ? $\rightarrow$ ask in \#ubuntu+1 $\rightarrow$ they are afk all of them $\rightarrow$ so ? doesn't mean its ontoppic here & \textbf{MrRNN Act.\@ -Ent.\@:}can you just tell me how ? \newline \textbf{MrRNN Noun:} join \#ubuntu-offtopic \newline \textbf{HRED + Act.\@-Ent.\@:} what are you trying to do ? \newline \textbf{HRED:} i know, but i dont know how to install it \newline \textbf{LSTM:} ? \\ \hline 
        How shall I reboot BIOS 100 time through script $\rightarrow$ i'm really curious why anyone would ever want to do that... $\rightarrow$ For BIOS stress testing & \textbf{MrRNN Act.\@ -Ent.\@:}do you know how to check bios ? \newline \textbf{MrRNN Noun:} How do I disable the bios ? \newline \newline \textbf{HRED + Act.\@-Ent.\@:} I don't want to reboot, I just want to be able to boot from a USB stick, but I don't want to have to do it every time I want to do it \newline \textbf{HRED:} that 's not what you want to do, but you might want to ask in \#\#hardware \newline \textbf{LSTM:} ? \\ \hline 
        what is the current version of wine in ubuntu ? i can see wine 1.5. However, when I try to install it via synaptic it wants to remove VLC ? $\rightarrow$ did you install vlc as a windows version ? $\rightarrow$ windows version on Ubuntu ? & \textbf{MrRNN Act.\@ -Ent.\@:}what version of ff do you have installed ? \newline \textbf{MrRNN Noun:} what version of ubuntu are you using ? \newline \textbf{HRED + Act.\@-Ent.\@:} how do i install the latest version of wine ? i have the latest version, but i dont know how to install it \newline \textbf{HRED:} yes \newline \textbf{LSTM:} yes \\ \hline 
        Hey guys, I'm trying to install libreoffice from the repositories, but it 's throwing a wobbly. Can anyone help ? Looks like dependency problems, but when I attempt to resolve them it throws a wobbly at me. \textless url \textgreater $\rightarrow$ try ' apt-get install -f ' yeah, it looks like you don't have java install You need to try to run ' apt-get install -f', just that string try ' apt-get purge libreoffice ' it looks like you had it installed previously $\rightarrow$ update : just realized I removed it earlier because it wasn't working. let me re-install. when I run ' apt-get purge libreoffice ' it just complains of unmet dependencies/ $\rightarrow$ but you were able to reinstall fine ? $\rightarrow$ no, it throws the errors seen in the pastebin at me $\rightarrow$ jwrigley, check to see if /var/lib/libreoffice/share/prereg exists $\rightarrow$ doesn't appear to exist when i cd to it in terminal $\rightarrow$ use apt-get via the command line tab completion helps a lot run ' apt-get purge libreoffice ' $\rightarrow$ oh, yeah i see now. sorry. now it is telling me to fix dependencies with apt-get install -f $\rightarrow$ awesome, so are you all set now ? $\rightarrow$ well libreoffice is still not working, so no. & \textbf{MrRNN Act.\@ -Ent.\@:} I see. \newline \textbf{MrRNN Noun:} how do I get rid of libreoffice ? \newline \textbf{HRED + Act.\@-Ent.\@:} it says " E : Could not get lock /var/lib/dpkg/lock - open ( 11 Resource temporarily unavailable ) E : Unable to locate package something " \newline \textbf{HRED:} I don't know what that is, but I don't know how to fix it \newline \textbf{LSTM:} is installed \\ \hline 
        Can anyone help get wireless working ? $\rightarrow$ whats the trouble ? $\rightarrow$ It doesnt show any wireless networks & \textbf{MrRNN Act.\@ -Ent.\@:} It 's not working. \newline \textbf{MrRNN Noun:} How do I get to the network manager ? \newline \textbf{HRED + Act.\@-Ent.\@:} what is the name of your wireless card ? \newline \textbf{HRED:} are you using ndiswrapper ? \newline \textbf{LSTM:} what do you mean ? \\ \hline 
        Who can answer : " How can i get mac address of the server i just connected via pppoe ?" $\rightarrow$ try tcpdump ? tcpdump -e & \textbf{MrRNN Act.\@ -Ent.\@:} what are you looking for ? \newline \textbf{MrRNN Noun:} what is your mac address ? \newline \textbf{HRED + Act.\@-Ent.\@:} what are you trying to do ? \newline \textbf{HRED:} thanks, i'll try that \newline \textbf{LSTM:} command not found \\ \hline
 \end{tabular}
\end{table}

\begin{table}[ht]
 \caption{Twitter model examples. The $\rightarrow$ token indicates a change of turn.}
 \label{table:twitter-examples}
 \scriptsize
 \centering
 \begin{tabular}{p{45mm}|p{80mm}}
 \textbf{Context} & \textbf{Response} \\ \hline \\ \hline 
     kids are so well informed! <url> & \textbf{MrRNN:} i love the kids!!!  \newline \textbf{VHRED:} wow, you're so cute! \newline \textbf{HRED:} i'm so happy for you!!! \newline \textbf{LSTM:} that's the spirit!!! \\ \hline 
     russell crowe. amazing actor and good person. perfect. i am sure. & \textbf{MrRNN:} johnny depp is the best!!!  \newline \textbf{VHRED:} something!!! \newline \textbf{HRED:} i'm glad you enjoyed it! \newline \textbf{LSTM:} he is a great actor. \\ \hline 
     thank you <name> for being so welcoming. who knew downtown little rock was so beautiful? loved it! & \textbf{MrRNN:} i love that song!!!  \newline \textbf{VHRED:} happy birthday to you! \newline \textbf{HRED:} i love you!!! \newline \textbf{LSTM:} i'm glad you liked it! \\ \hline 
     i've got a core i5 macbook air right here if anyone has any questions. or you can ask in the comments thread here <url> $\rightarrow$ how well does it run photoshop illustrator? $\rightarrow$ pretty much just as well as the current macbook pros do. & \textbf{MrRNN:} do you have a macbook pro???!!!  \newline \textbf{VHRED:} ok, so what do you want to do? \newline \textbf{HRED:} i'm sure you'll find it soon. i'm sure it'll be worth it. \newline \textbf{LSTM:} thanks for the heads up. i'll check it out.  \\ \hline 
     maybe bigger than jb. haha jk. $\rightarrow$ i'm fan of her. she's so awesome. $\rightarrow$ hahahahahahahaha p just amazed. $\rightarrow$ thinking of dudz? & \textbf{MrRNN:} yes. i love justin bieber!!!  \newline \textbf{VHRED:} i love her. \newline \textbf{HRED:} no, i don 't think so. \newline \textbf{LSTM:} i think she's just a little bit of a something.  \\ \hline 
     <number> countdown reminds me of hold on wilsonphillips & \textbf{MrRNN:} it's a spring break!!! - - - - - - - - - - - - - - - - -  \newline \textbf{VHRED:} i'm so excited for you!!! \newline \textbf{HRED:} something!!! \newline \textbf{LSTM:} something!!!  \\ \hline 
     here is a sadbuttrue fact - bad news travels fast while good news seems to just linger. people don 't like spreading good news & \textbf{MrRNN:} i agree with the news!!!  \newline \textbf{VHRED:} thank you for the support! \newline \textbf{HRED:} what's the news? \newline \textbf{LSTM:} i agree, but i don 't think it's a bad thing to do. i don 't think it's a bad thing to do.  \\ \hline 
     have you dmed justin bieber yet for me? <smiley> $\rightarrow$ yeah. & \textbf{MrRNN:} do you have bieber fever? <smiley>  \newline \textbf{VHRED:} dm me your email and i'll send it to you \newline \textbf{HRED:} thank you \newline \textbf{LSTM:} thank you <smiley>  \\ \hline 
     hanging out in the library for the past couple hours makes me feel like i'll do great on this test! $\rightarrow$ wow, what a nerd lol jk haha p what!? you changed your bio $\rightarrow$ do you like my bio now? i feel bad for changing it but i like change. p & \textbf{MrRNN:} yeah, it's in the library!!!  \newline \textbf{VHRED:} what do you mean? \newline \textbf{HRED:} haha i'm glad i'm not the only one! \newline \textbf{LSTM:} yeah i like it but i don 't know what to do with it lol  \\ \hline 
     christian bale must be prepping for a gordon's fish sticks commercial. oscar & \textbf{MrRNN:} i'm not a christian bale. i don 't know what he's talking about. but i don 't know what he's talking about. lol. something!!!  \newline \textbf{VHRED:} did you see the new harry potter movie? \newline \textbf{HRED:} i don 't think it's a good movie, but it's a good movie. \newline \textbf{LSTM:} i'm going to have to watch that.  \\ \hline 
     some railscamp nz action. <url> $\rightarrow$ how goes it? & \textbf{MrRNN:} it's great to see you in nz!!!  \newline \textbf{VHRED:} not too bad, just a little tired. how are you? \newline \textbf{HRED:} it's been a while since i've been here. \newline \textbf{HRED:} it's going to be a long day.  \\ \hline 
     second netball game for the week. let's do it! isupportwomeninsport & \textbf{MrRNN:} what's netball???!!!  \newline \textbf{VHRED:} i'm going to have to go to the gym tomorrow. \newline \textbf{HRED:} i'm going to be in the <number>'s. \newline \textbf{LSTM:} what's the score? \\ \hline
 \end{tabular}
\end{table}